\documentclass[journal, twoside]{IEEEtran}
\let\labelindent\relax
\usepackage{amsmath}
\usepackage{amssymb}
\usepackage{mathtools}
\usepackage{tabularx}
\usepackage{ulem}
\usepackage{threeparttable}
\usepackage{booktabs}
\usepackage{graphicx}
\usepackage{wrapfig}
\usepackage[lined, ruled, linesnumbered, commentsnumbered, noend]{algorithm2e}
\usepackage[colorlinks=true, allcolors=blue]{hyperref}

\SetCommentSty{mycommfont}
\usepackage{subcaption}
\usepackage{eqlist}
\usepackage{amsfonts}
\usepackage{enumitem}
\usepackage{lipsum}
\usepackage{stfloats}
\usepackage[numbers]{natbib}
\usepackage{xcolor}
\usepackage{comment}
\usepackage{multirow}

\newcommand{\rev}[1]{\texorpdfstring{\begingroup\color{black}#1\endgroup}{#1}}
\newcommand{\limebox}[1]{\setlength{\fboxsep}{0pt}{\colorbox{lime}{#1}}}
\newcommand{\pinkbox}[1]{\setlength{\fboxsep}{0pt}{\colorbox{pink}{#1}}}

\begin{document}
% \includepdf[pages=-]{response.pdf}
\title{Learning to Push, Group, and Grasp: A Diffusion Policy Approach for Multi-Object Delivery}
% two-dimensional manifolds
\author{Takahiro Yonemaru, Weiwei Wan$^{*}$, Tatsuki Nishimura, Kensuke Harada
\thanks{All authors are from the Graduate School of Engineering Science, Osaka University, Japan. \\Contact: Weiwei Wan, {\tt\small wan@sys.es.osaka-u.ac.jp}}}
\markboth{Under review by a robotics journal}
{Yonemaru \MakeLowercase{\textit{et al.}}: Learning to Push, Group, and Grasp: A Diffusion Policy Approach for Multi-Object Delivery}
\maketitle

% \author{Authors' names have been anonymized following the IEEE RAS Double-Anonymous Review Process Guidelines
% \thanks{Authors' affiliations and contact information have been anonymized following the IEEE RAS Double-Anonymous Review Process Guidelines.}}
% \markboth{IEEE Robotics and Automation Letters. Re-Submission for Review, 2025.}
% {Authors' names anonymized: Learning to Group and Grasp Multiple Objects}
% \maketitle

%%%%%%%%%%%%%%%%%%%%%%%%%%%%%%%%%%%%%%%%%%%%%%%%%%%%%%%%%%%%%%%%%%%%%%%%%%%%%%%%
\begin{abstract}
Simultaneously grasping and delivering multiple objects can significantly enhance robotic work efficiency and has been a key research focus for decades. The primary challenge lies in determining how to push objects, group them, and execute simultaneous grasping for respective groups while considering object distribution and the hardware constraints of the robot. Traditional rule-based methods struggle to flexibly adapt to diverse scenarios. To address this challenge, this paper proposes an imitation learning-based approach. We collect a series of expert demonstrations through teleoperation and train a diffusion policy network, enabling the robot to dynamically generate action sequences for pushing, grouping, and grasping, thereby facilitating efficient multi-object grasping and delivery. We conducted experiments to evaluate the method under different training dataset sizes, varying object quantities, and real-world object scenarios. The results demonstrate that the proposed approach can effectively and adaptively generate multi-object grouping and grasping strategies. With the support of more training data, imitation learning is expected to be an effective approach for solving the multi-object grasping problem.
\end{abstract}

\begin{IEEEkeywords}
Multi-Object Grasping, Imitation Learning
\end{IEEEkeywords}

%==============================================================================================================================
\section{Introduction} % 1.5 page

% \IEEEPARstart{I}{n} recent years, advances in AI and sensor technology have enabled the development of robots capable of performing picking tasks in environments where objects are randomly scattered. In such situations, robots need to deliver objects efficiently to their target locations. However, a common limitation of robots is that they can typically handle only one object at a time, which leads to slower task execution. Different from robots, humans can grasp and carry multiple objects at once, and are thus more efficient. To bridge this gap, this study focuses on enabling robots to perform simultaneous multi-object grasping, thereby improving their work efficiency.
\IEEEPARstart{T}{his} study focuses on enabling robots to perform simultaneous multi-object grasping while considering pushing-based grouping. Previous studies on simultaneous multi-object grasping have identified several challenges. For example, Sakamoto et al. \cite{sakamoto2021efficient}  proposed a rule-based method to push objects closer together before grasping them simultaneously. However, the method cannot handle complex pushing actions, and the combinations of objects that can be grasped together are limited. Agboh et al. \cite{agboh2022multi} studied identifying near objects arranged on a plane and grasping multiple objects by pushing them together as the gripper closes. While effective under specific conditions, the method assumes that objects are initially spaced within the gripper’s maximum opening width. Or else, it degenerates into single-object grasping. These limitations highlight the need for more advanced strategies that can group objects through flexible pushing actions and select suitable object combinations for simultaneous grasping. Designing such strategies through hand-crafted rules, however, remains a challenging task.

\begin{figure}[tbp]
    \centering
    \includegraphics[width=\linewidth]{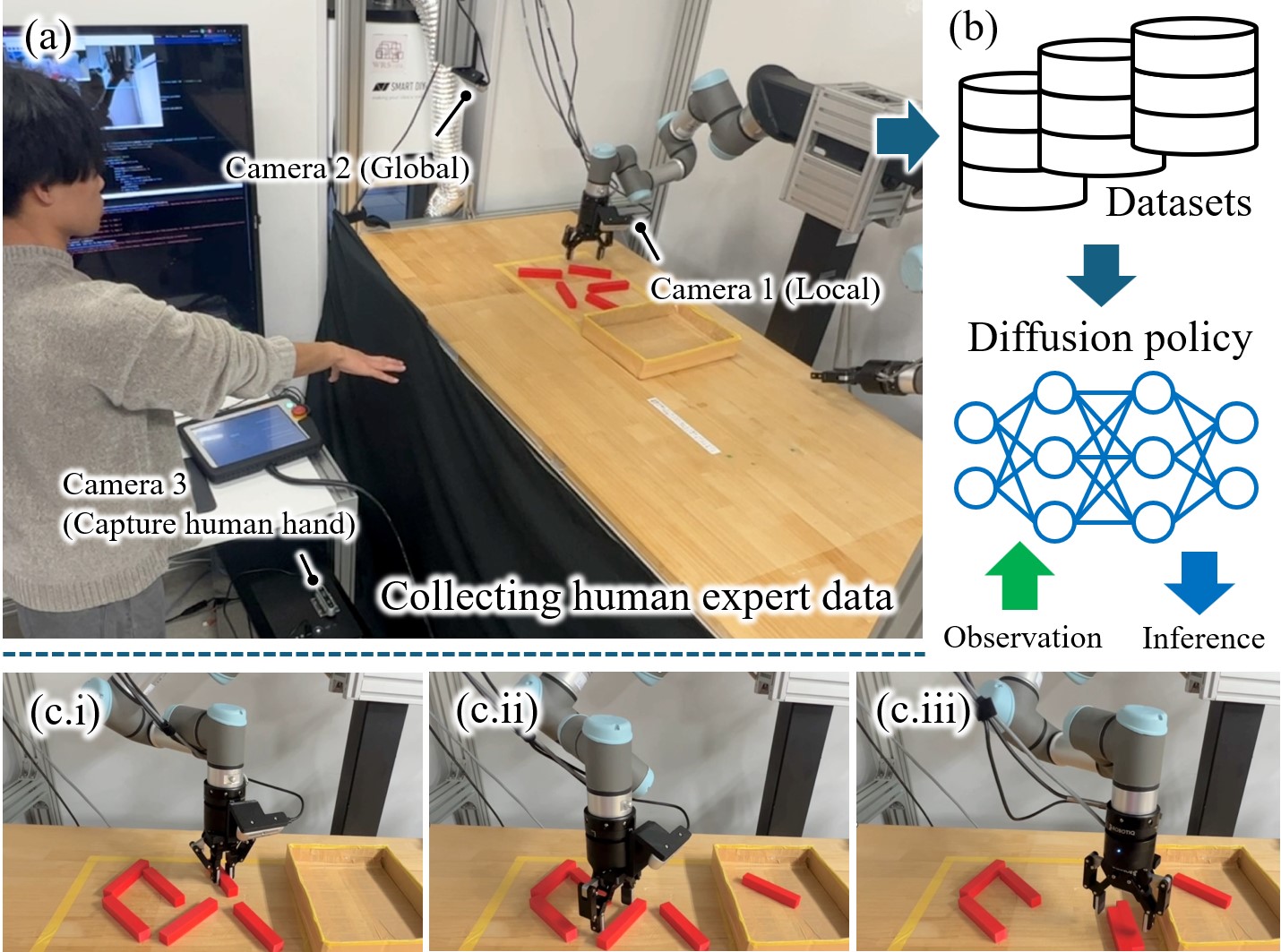}
    \caption{(a) Collecting human expert data using teleoperation. (b) The collected dataset is used to train a diffusion policy network for inferring actions. (c) The policy network uses past observations to generate pushing or grasping actions.}
    \label{workflow}
\end{figure}  

In this context, we propose addressing these issues by using imitation learning and enabling robots to learn how humans decide which objects to group and grasp. By acquiring grasping strategies through imitation learning, the proposed method enables robots to automatically choose suitable object combinations and perform the necessary pushing actions to group them. The method can overcome the limitations of existing methods and improve the efficiency and versatility of robotic picking tasks. Fig. \ref{workflow} illustrates the components and workflow of the proposed framework. In the data collection phase shown in Fig. \ref{workflow}(a), a teleoperation system is developed for human operators to control the robot using natural hand movements. The setup enables collecting data like robot joint angles, gripper opening and closing states, and images from the global and local cameras mounted in the environment and on the robot hand. In the learning phase shown in Fig. \ref{workflow}(b), the framework employs Diffusion Policy \cite{chi2023diffusion} to learn appropriate robot actions several steps ahead, based on the robot's state from previous steps. Finally, in the inference phase shown in Fig. \ref{workflow}(c), the framework uses the robot's gripper poses, opening states, and camera images observed from a few steps before to predict new gripper poses and states for subsequent steps, thereby generating pushing and simultaneous grasping motions for multiple objects. 

\rev{Four} groups of experiments were conducted to validate the proposed framework. The first group focused on comparing and analyzing the influence of different training data volumes and object quantities on learning and inference. \rev{The second group conducted a comparative study against a rule-based analytical method.} The \rev{third} group conducted fine-tuning of \rev{models} pre-trained on a dataset of simple objects to more practical scenarios and evaluated the generality of the proposed method to real-world objects. The \rev{fourth} group evaluates the generalization ability of the learned policy across different object geometries. The experimental results validated the effectiveness of the proposed method and the adaptivity of fine-tuned models. Additionally, the results indicated that training a well-performing policy network requires a large amount of high-quality data, with a positive correlation between data volume and learning performance. Developing efficient methods for collecting such data could potentially improve both learning and inference performance.

To the best of our knowledge, this is the first work to apply Diffusion Policy to choosing among pushing, grouping, and multi-object grasping. The proposed method is simple to implement and data-driven. We provide extensive experiments and in-depth analysis, and discuss both the strengths and limitations of the method. With larger-scale training data, imitation learning is expected to become an even more effective solution for multi-object grasping. We hope our findings and insights can inspire further research in this direction.

\section{Related Work} % 2 page

Multi-object simultaneous grasping has been attracting significant attention for decades. As seminal work, Aiyama et al. \cite{aiyama98} studied using two manipulators to cooperatively grasp and lift boxes. Harada and Kaneko \cite{harada00neighbouring}\cite{harada98enveloping} studied the neighboring equilibrium of multiple cylindrical objects and their enveloping grasps. Yoshikawa et al. \cite{yoshikawa2001optimization} proposed the optimal power grasp condition for multiple objects from the viewpoint of reducing finger joint torques. Yamada et al. \cite{yamada2009grasp}\cite{yamada2012stability}\cite{yamada2015static} analyzed the grasp stability of multiple objects by considering object surface curvatures and contact spring models. 
% The authors also extended the theories to multi-fingered hands \cite{yamada2012stability} and 3D shapes \cite{yamada2015static}. 

More recent studies by Chen et al. \cite{chen2021multiobjectgrasping} inserted the Barrett hand into a pile of objects and estimated the number of grasped items using a deep learning model. Shenoy et al. \cite{shenoy2022multi} focused on efficiently transferring the grasped objects to a separate container. Takahashi et al. \cite{takahashi2021uncertainty} studied grasping a given amount of granular foods. Post-grasping methods were explored for fine adjustments \cite{takahashi2021target}. Li et al. \cite{li2024graspmultiple} developed a framework for planning and learning multi-object grasping (lifting). Jiang et al. \cite{jiang24multiobject} developed a vacuum gripper equipped with multiple suction cups and the related grasp planning methods to simultaneously pick multiple objects from a tray. Nguyen et al. \cite{nguyen2023wiringclaw} proposed the Wiring-Claw Gripper, which took advantage of the soft interaction between wires in a claw and objects for simultaneous multi-object grasping. Yao et al. \cite{yao2023exploiting} used a dexterous hand with progressive planning to exploit redundancy for multi-object grasping.

The above approaches could achieve stable grasping of multiple or unknown objects. However, they often assumed that the objects were initially positioned in proximity and were less applicable to scattered scenes. To tackle such limitations, several studies have explored using pushing actions to rearrange objects before simultaneous grasping. For example, Sakamoto et al. \cite{sakamoto2021efficient} proposed a method that can determine whether to grasp a single object, grasp two objects simultaneously, or push one object closer for simultaneous grasping based on the distance and friction between objects. Agboh et al. \cite{agboh2022multi}\cite{agboh2023learning} proposed the $\mu$-MOG framework to simultaneously grasp multiple rigid convex polygonal objects scattered on a plane. The method took advantage of a gripper's opening jaw to align objects before grasping. Shrey et al. \cite{aeron2023push} leveraged linear pushing actions along lines perpendicular to the gripper fingers to bring object centroids closer, allowing a robot to grasp multiple objects that initially lay beyond the gripper's opening width. Ye et al. \cite{ye2023pickmultiobjectpicking} used weighted graphs to model object distributions and assess collision-free clusters, thereby enabling efficient and reliable multi-object picking. Kishore et al. \cite{srinivas2023busboy} proposed a method for efficiently delivering cups, bowls, and utensils on tables by combining pushing and stacking actions. The method especially employed inward-pushing to reduce positional uncertainty. 

The pushing-based methods have addressed the problem of scattered objects to some extent. However, they are predominantly rule-based, making it challenging to generate complex pushing trajectories for intricate grouping and grasping. Previously, reinforcement learning methods have attracted much attention in robotic pushing and grasping as they can adapt to various object arrangements. For instance, Zeng et al. \cite{zeng2018learning} achieved efficient grasping by enabling the complementary relationship between pushing and grasping actions to be learned through self-supervised deep reinforcement learning. Chen et al. \cite{chen2020combining} proposed a grasping strategy that combines rule-based methods with reinforcement learning to integrate pushing and grasping. Wang et al. \cite{wang2024self} attained high success rates and robustness by leveraging self-supervised deep reinforcement learning to learn an integrated model of pushing and grasping. These studies demonstrate that reinforcement learning has the advantage of adapting to unknown environments. However, a key challenge in reinforcement learning lies in the design of rewards. This issue becomes particularly pronounced in the context of this study as we use pushing to achieve grouping, followed by grasping. Designing rewards that facilitate rapid convergence for the pushing-grouping-grasping routine is notably difficult. A learning method conceptually similar to reinforcement learning but not dependent on reward design is imitation learning. Recent studies have showcased the potential of imitation learning in robotic manipulation applications. For example, ACT \cite{zhao2023learning} has achieved high success rates in delicate tasks such as opening and closing Ziploc bags and inserting batteries into controllers. Diffusion policy \cite{chi2023diffusion} has enabled high success rates in tasks requiring diverse actions and significant flexibility, such as spreading sauce on pizza dough or relocating T-shape blocks and mugs placed at random positions on a plane to target positions and orientations. Given its advantages, this research adopts imitation learning as the primary methodology. By learning flexible pushing actions and grasping decisions from human demonstrations, our approach enables dynamic selection of object combinations, rearrangement through complex pushing trajectories, and adaptive grasping sequences that were previously difficult to attain with rule-based or reinforcement learning methods.

\section{Data Collection Using Visual Teleoperation}

\begin{figure}[!t]
    \centering
    \includegraphics[width=\linewidth]{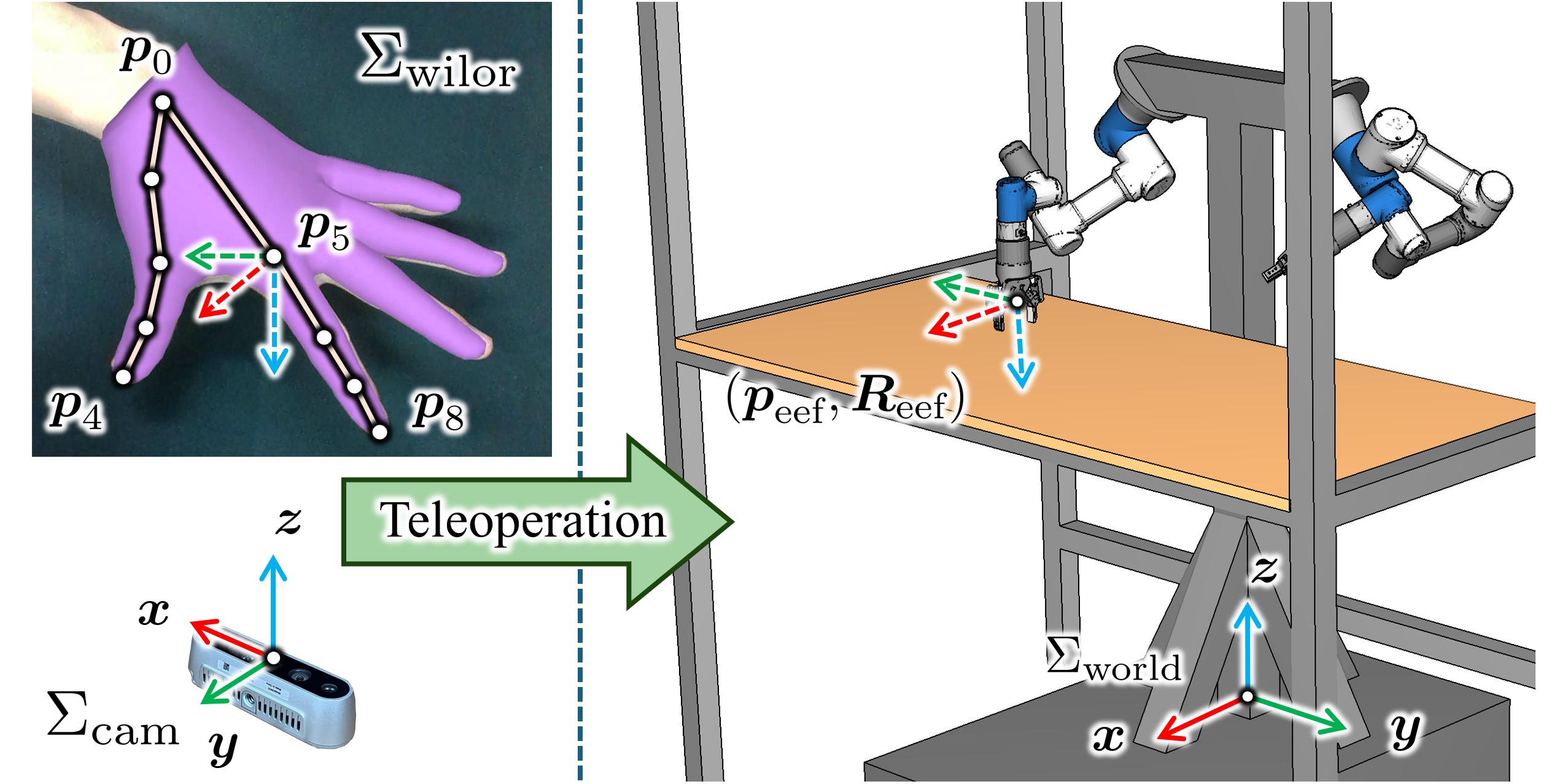}
    \caption{Mapping relationship between hand poses detected by WiLoR and the end-effector of the target robot.} 
    \label{mapping}
\end{figure}

Similar to AnyTeleop \cite{qin2023anyteleop}, we develop a teleoperation system for a robot by detecting the hand skeleton pose using an RGB-D camera. We use WiLoR \cite{potamias2024wilor} for detecting the hand skeleton pose. The output of WiLoR is a 3D skeleton described in the $\Sigma_\text{wl}$ coordinate system shown in Fig. \ref{mapping}. It is then integrated with depth values of the same RGB-D camera to derive the 3D skeleton in the coordinate system of the depth camera, $\Sigma_\textnormal{cam}$. After that, we transform the 3D data into actions for the robot, thereby enabling teleoperation. The coordinates and transformations used in the process are as follows.

\textit{Jaw bottom of the robot gripper:} 
The jaw bottom center of the robot's end-effector is defined as a position corresponding to the MCP joint of the index finger ($\boldsymbol{p}_{5}$ in Fig. \ref{mapping}) and is transformed to the world coordinate system using
\begin{equation}
{}^{\Sigma_\text{wd}}\boldsymbol{p}_{\text{eef}} = {}^{\Sigma_\text{wd}}\mathbf{R}_{\Sigma_\text{cam}} \cdot
\left(
{}^{\Sigma_\text{cam}}\mathbf{R}_{\Sigma_\text{wl}} \cdot {}^{\Sigma_\text{wl}}\boldsymbol{p}_{5}
+
{}^{\Sigma_\text{cam}}\mathbf{p}_{\text{wl}}
\right)\nonumber +
{}^{\Sigma_\text{wd}}\mathbf{p}_{\text{cam}}
\label{affine-pos}
\nonumber
\end{equation}

\textit{Rotation of the robot gripper:} 
The rotation of the end-effector, ${}^{\Sigma_\text{wd}}\boldsymbol{R}_{\textnormal{eef}}$, is constrained such that its approaching direction is always vertically downward (the last column or $\boldsymbol{z}$ axis of ${}^{\Sigma_\text{wd}}\boldsymbol{R}_{\textnormal{eef}}$ is parallel with and inversely aligns with the $\boldsymbol{z}$ axis of $\Sigma_{\textnormal{world}}$ in Fig. \ref{mapping}(a)). The rotation around the approaching direction is defined using the position of the thumb tip (${}^{\Sigma_\text{wl}}\boldsymbol{p}_4$) and the position of the index finger tip (${}^{\Sigma_\text{wl}}\boldsymbol{p}_8$) following equation \eqref{ee_rot}.
\begin{equation}
{}^{\Sigma_\text{wd}}\boldsymbol{R}_\text{eef}
=\begin{bmatrix}\hat{\boldsymbol{d}}_x & \hat{\boldsymbol{d}}_y & 0 \\\hat{\boldsymbol{d}}_y &-\hat{\boldsymbol{d}}_x & 0 \\0 & 0 & 1 
\end{bmatrix},~
\hat{\boldsymbol{d}} = \frac{{}^{\Sigma_\text{wl}}\boldsymbol{p}_8 - {}^{\Sigma_\text{wl}}\boldsymbol{p}_4}{\|{}^{\Sigma_\text{wl}}\boldsymbol{p}_8 - {}^{\Sigma_\text{wl}}\boldsymbol{p}_4\|}
\nonumber
\label{ee_rot}
\end{equation}

\textit{Gripping state:} The robot gripper's open/close state, $g$, is defined based on the distance between the thumb and the index finger following
\begin{equation}
\label{gripperstate}
g = 
\begin{cases}
0 & \text{if } d \geq \mu_\text{up} \\
1 & \text{if } d < \mu_\text{bot}
\end{cases},~
d =\frac{\|{}^{\Sigma_\text{wl}}\boldsymbol{p}_8 - {}^{\Sigma_\text{wl}}\boldsymbol{p}_4\| }{\max\|{}^{\Sigma_\text{wl}}\boldsymbol{p}_8 - {}^{\Sigma_\text{wl}}\boldsymbol{p}_4\|}
\nonumber
\end{equation}
Here, $g=0$ denotes the gripper in an open state, while $g=1$ indicates the gripper in a closed state. The threshold values $\mu_\text{up}$ and $\mu_\text{up}$ are chosen to be 0.8 and 0.1 for practical purpose. The distance between the thumb and the index finger is normalized by dividing its maximum distance.

Based on these mapping relationships, we define the observation $\boldsymbol{o}_{t}$ at a timestamp $t$ as $\boldsymbol{o}_{t} = \{\boldsymbol{i}^{\text{gcam}}_t,\boldsymbol{i}^{\text{lcam}}_t,\boldsymbol{s}_{t}\}$, where $\boldsymbol{i}^{\text{gcam}}_t, \boldsymbol{i}^{\text{lcam}}_t$ are video frames captured by the global and local cameras. $\boldsymbol{s}_t$ is the state of the robot and is defined as
\begin{equation}
    \boldsymbol{s}_t=\left\{{p}_x,{p}_y,{p}_z,{q}_w,{q}_x,{q}_y,{q}_z,{g}\right\}.
    \nonumber
\end{equation}
Here, $({p}_x,{p}_y,{p}_z)$ are the ${}^{\Sigma_\text{wd}}\boldsymbol{p}_{\text{eef}}$ position of the robot at time $t$. $({q}_w,{q}_x,{q}_y,{q}_z)$ are the ${}^{\Sigma_\text{wd}}\textbf{R}_{\text{eef}}$ rotation of the robot at time $t$. $g$ is the gripper open/close state at time $t$.

Based on this observation and the definition of the robot's state, the robot's action at timestamp $t$ is defined as:
\begin{equation}
    \boldsymbol{a}_t=\left\{{p}_x,{p}_y,{p}_z,{q}_w,{q}_x,{q}_y,{q}_z, {g}\right\}.
\end{equation}
It has the same mathematical representation as $\boldsymbol{s}_t$, except that the constituent values denote the position, rotation, and gripper state that the robot is going to act on at the next time step.We clarified the above discussion at the beginning of the experimental section.

During data collection, a human operator controls the robot via the teleoperation system to demonstrate how to group and grasp multiple objects. The collected demonstrations were used to train a Diffusion Policy to learn human strategies. Further details are provided in the following section\footnote{A follow-up interview with the human operator revealed the overall heuristics used during teleoperated data collection. To avoid overloading the gripper, the number of grouped objects was limited to three, based on object size. Objects were grouped if they were spatially close and roughly aligned to avoid significant rotational adjustments. When choosing the first object, the operator tended to select one surrounded by nearby candidates to facilitate multi-object grouping. These human strategies may offer useful inspiration for readers interested in understanding the underlying rules.}.

\section{Network Selection and Learning}

\begin{figure}[!t]
    \centering
    \includegraphics[width=\linewidth]{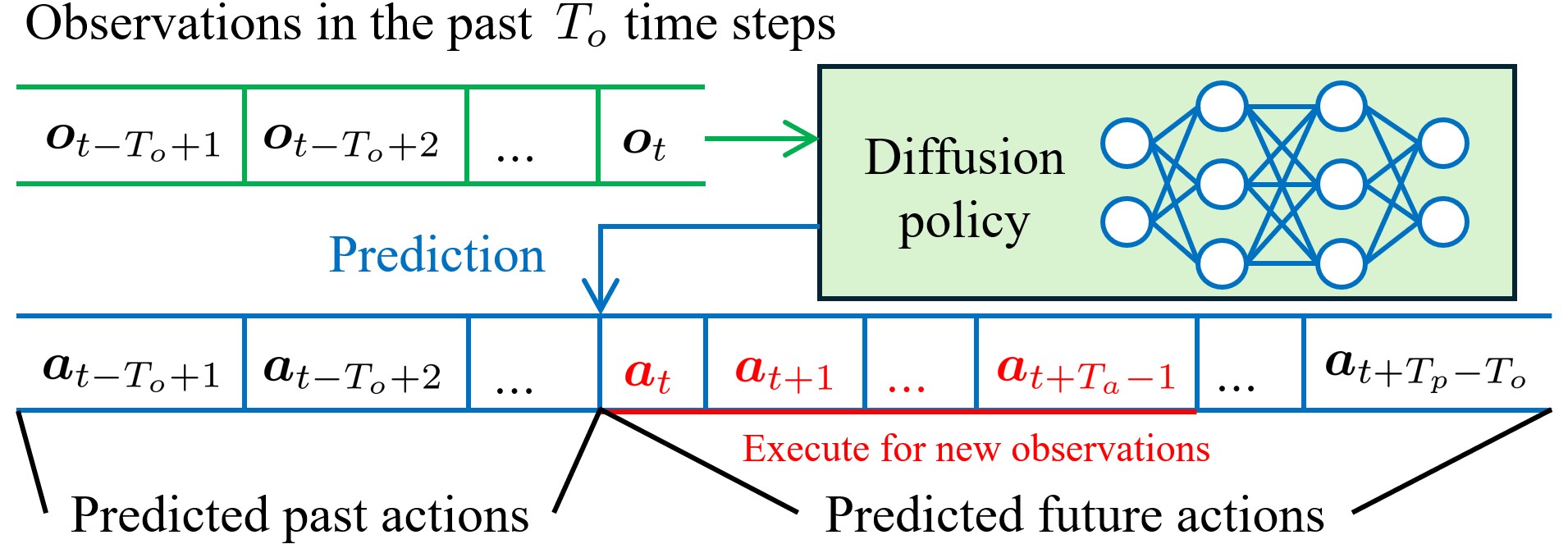}
    \caption{The diffusion policy takes as input the observations from the past $T_o$ time steps. It infers likely actions for the past $T_o$ time steps and predicts actions for the future $T_p-T_o$ time steps. The first $T_a$ predicted future actions are executed for iterative observation-prediction looping.}
    \label{network_io}
\end{figure}

Based on the past $T_o$ time steps of observations at time $t$ $\boldsymbol{O}_{t}=\{\boldsymbol{o}_{t-T_o+1}$, $\boldsymbol{o}_{t-T_o+2}$, $...$, $\boldsymbol{o}_t\}$, our diffusion policy is expect to infer actions for $T_{p}$ time steps, namely $\boldsymbol{A}_{t}=\{\boldsymbol{a}_{t-T_o+1}$, $\boldsymbol{a}_{t-T_o+2}$, $...$, $\boldsymbol{a}_t$, $\boldsymbol{a}_{t+1}$, $...$, $\boldsymbol{a}_{t+T_p-T_o}\}$. The inferred $\boldsymbol{A}_{t}$ includes the predicted past actions $\{\boldsymbol{a}_{t-T_o+1}$, $\boldsymbol{a}_{t-T_o+2}$, $...$, $\boldsymbol{a}_{t-1}\}$ and the predicted future actions $\{\boldsymbol{a}_{t}$, $\boldsymbol{a}_{t+1}$, $...$, $\boldsymbol{a}_{t+T_p-T_o}\}$. The predicted past actions correspond to the observations of the past $T_o$ time steps. The predicted future actions are the learned policies for the next $T_p-T_o$ time steps. The robot will execute $T_a$ steps of actions $\{\boldsymbol{a}_{t}, \boldsymbol{a}_{t+1}, ..., \boldsymbol{a}_{t+T_a-1}\}$ for collecting new observations and conduct new predictions. $T_a$ is selected to be a value smaller than $T_p-T_o$. Fig. \ref{network_io} illustrates the details of the time step division for observation, prediction, and execution.

\begin{figure}[!t]
    \centering
    \includegraphics[width=\linewidth]{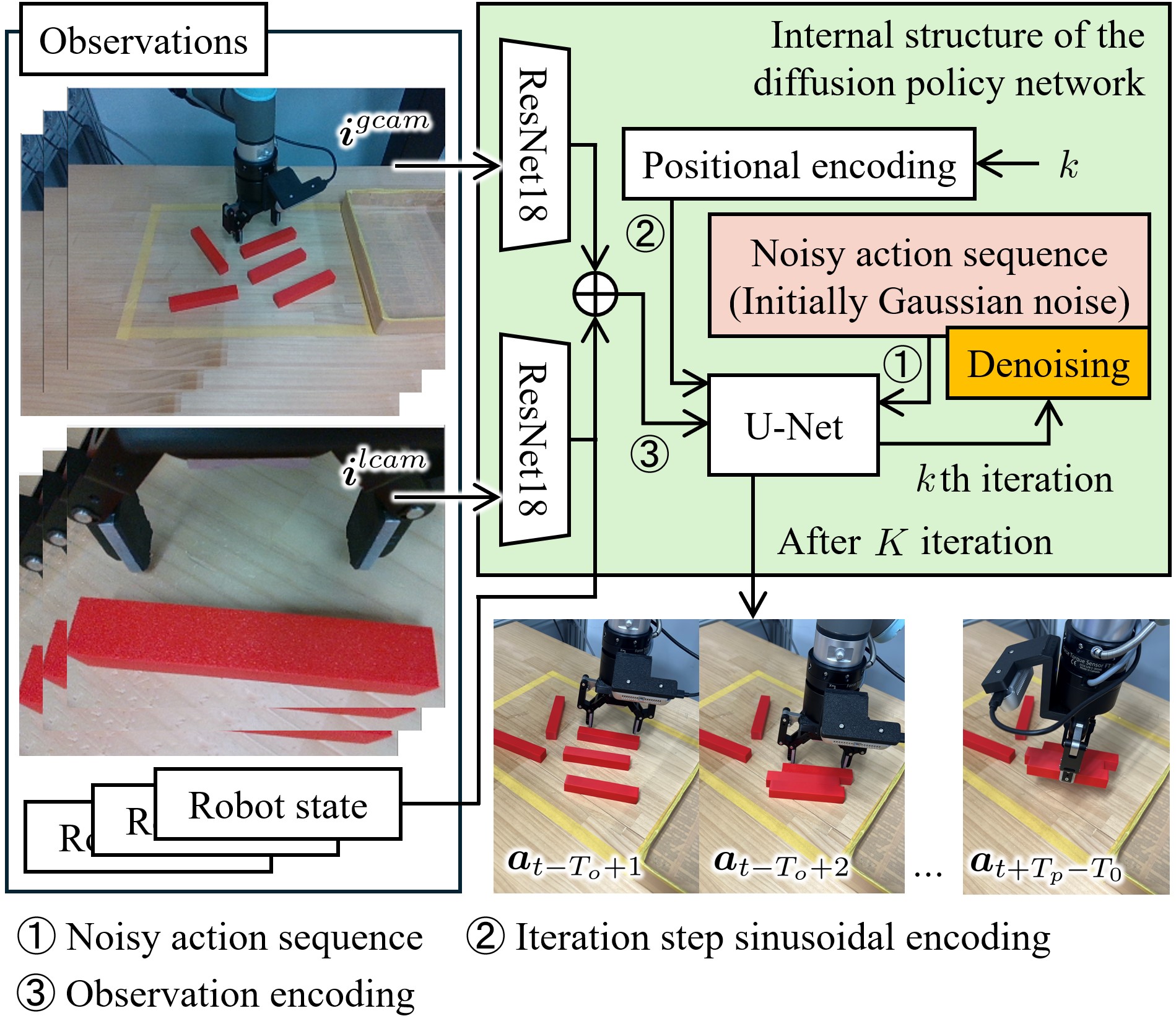}
    \caption{Internal mechanism of the diffusion policy network. It has a U-Net core that accepts a noisy action sequence and predicts the noise within it. Through $K$ iterations, the U-Net core progressively removes the noise to acquire the action sequence most suitable for the observations.}
    \label{denoising}
\end{figure}

The internal structure of the used method is illustrated in Fig. \ref{denoising}. There are in total $T_o$ observations. Each observation consists of two image frames respectively captured from the global and local cameras. They are processed separately using two ResNet-18 for feature extraction. Each image frame is compressed into a $N$-dimensional feature vector after feature extraction, and the total observations form two $N\times T_o$-dimensional vectors. Additionally, the observation includes the robot's state, which is an $8$-dimensional vector for each observation as discussed in the previous section. In total, the robot states form a $8\times T_o$-dimensional vector. The two $N\times T_o$-dimensional vectors together with the $8\times T_o$-dimensional robot state vector are concatenated to form a $(2N+8)\times T_o$-dimensional vector which serves as the observation embedding of a U-Net core used in the diffusion policy network. 

The U-Net core takes a noisy action sequence (represented as a $8\times T_p$-dimensional vector) as input and predicts the noise in the representation. The output of the U-Net is another $8\times T_p$-dimensional vector where each value corresponds to the estimated noise in the input sequence. This predicted noise is subsequently subtracted from the input sequence within the denoising box shown in Fig. \ref{denoising}, which is mathematically
\begin{equation}
\boldsymbol{A}_{t}^{k-1}=\alpha(\boldsymbol{A}_{t}^{k}-\gamma{\boldsymbol{\varepsilon_\theta}}(\boldsymbol{A}_{t}^{k}, \boldsymbol{O}_{t}, k)+\mathcal{N}(0, \sigma^2\boldsymbol{I})).
\end{equation}
Here, $k$ indicates the $k$th denoising step. $\boldsymbol{A}_{t}^{k}$ represents the noisy action sequence at the $k$th denoising step. $\boldsymbol{\varepsilon_\theta}()$ represents the noise prediction network (namely the U-Net core). $\theta$ indicates the parameters of the network. For each denoising iteration, the iteration step index $k$ is encoded using a $P$-dimensional sinusoidal positional embedding and is also incorporated into the U-Net core to provide temporal context. The positional embedding helps guide the network in estimating how much noise remains to be removed. $\alpha$, $\gamma$, $\sigma$ are parameters for controlling the learning stability\footnote{The default values from the LeRobot source code was adopted. Check the following webpage for details. https://github.com/huggingface/lerobot}. The noisy action sequence is initialized by sampling a Gaussian noise distribution. After undergoing $K$ iterative denoising steps, the diffusion policy network will output the denoised $\boldsymbol{A}_{t}^{0}$ as predicted action sequences.

To train the denoising network, we draw a sequence of $\boldsymbol{O}_{t}$ and a sequence of expert action $\boldsymbol{A}_{t}^0$ from the collected data set, select a random diffusion step $k$ and add Gaussian noises $\boldsymbol{\varepsilon}^k$ to the expert actions. The parameters $\theta$ of the U-Net core are learned by minimizing 
\begin{equation}
    \mathcal{L}=||\boldsymbol{\varepsilon}^k-\boldsymbol{\varepsilon_\theta}(\boldsymbol{A}_{t}^0+\boldsymbol{\varepsilon}^k, \boldsymbol{O}_{t}, k)||^2.
\end{equation}

\begin{table*}[!t]
\setlength{\tabcolsep}{2.5pt}
% \captionsetup{labelfont={color=magenta}, textfont={color=magenta}}
\caption{Results of randomly placed 3D-printed rectangular cuboids}
\label{tab_exp}
\centering
\rev{
\begin{threeparttable}
\begin{tabular}{c|cclcc|cclcc|cclcc|cclcc|cclcc|cclcc}
\toprule
\multirow{4.5}{*}{ID} 
& \multicolumn{10}{c|}{Data = 100} 
& \multicolumn{10}{c|}{Data = 200}  
& \multicolumn{10}{c}{Data = 300} \\ 
\cmidrule(lr){2-11} \cmidrule(lr){12-21} \cmidrule(lr){22-31}
& \multicolumn{5}{c|}{\# Objects: 4} 
& \multicolumn{5}{c|}{\# Objects: 6} 
& \multicolumn{5}{c|}{\# Objects: 4} 
& \multicolumn{5}{c|}{\# Objects: 6} 
& \multicolumn{5}{c|}{\# Objects: 4} 
& \multicolumn{5}{c}{\# Objects: 6}\\ 
\cmidrule(lr){2-6} \cmidrule(lr){7-11} \cmidrule(lr){12-16} \cmidrule(lr){17-21} \cmidrule(lr){22-26} \cmidrule(lr){27-31}
& \#P & \#G & Seq. & C & R
& \#P & \#G & Seq. & C & R
& \#P & \#G & Seq. & C & R
& \#P & \#G & Seq. & C & R
& \#P & \#G & Seq. & C & R
& \#P & \#G & Seq. & C & R \\
\midrule
1  & 2 &2 &{(1,2)} &N & P & 0 &0 &- &N & P& 1 &1 &{(2)} &N & P& 3 &3 &{(3,2,1)} &Y & - &2 &2 &(2,2)&Y &- &3&3&(3,1,2)&Y&-\\
2  & 3 &3 &{(1,1,2)} &Y & - & 0 &0 &- &N & P& 2 &2 &{(3,1)} &Y & -& 4 &4 &{(1,2,2,1)} &Y & -&3&3 &(1,2,1)&Y & -&4&4&(2,1,1,2)&Y&-\\
3  & 0 &0 &- &N & P& 0 &0 &- &N & P& 3 &3 &{(2,1,1)} &Y & -& 3 &3 &{(2,3,1)} &Y & -&3&3 &(1,2,1)&Y & -&4&4&(2,1,2,1)&Y&-\\
4  & 0 &0 &- &N & P& 0 &0 &- &N & P& 3 &3 &{(2,1,1)} &Y & -& 3 &2 &{(1,3)} &N&G&2&2 &(1,3)&Y & -&1&0& -&N&G\\
5  & 1 &0 &- &N&G& 1 &1 &{(1)} &N & P& 2 &2 &{(2,2)} &Y & -& 3 &3 &{(3,2,1)} &Y & -&2&2 &(3,1)&Y & -&1&1&(2)&N&P\\
6  & 2 &2 &{(2,2)} &Y & -& 0 &0 &- &N & P& 3 &3 &{(1,2,1)} &Y & -& 4 &4 &{(1,2,1,2)} &Y & -&3&3 &(1,2,1)&Y & -&4&4&(1,1,3,1)&Y&-\\
7  & 2 &2 &{(2,2)} &Y & -& 2 &2 &{(1,2)} &N & P& 1 &1 &{(3)} &N & P& 5 &5 &{(2,1,1,1,1)} &Y & -&3&3 &(2,1,1)&Y & -&3&3&(2,1,3)&Y&-\\
8  & 1 &1 &{(1)} &N & P& 2 &2 &{(1,1)} &N & P& 4 &4 &{(1,1,1,1)} &Y & -& 1 &1 &{(3)} &N & P&0&0 &-&N & P&3&3&(2,2,2)&Y&-\\
9  & 2 &2 &{(2,2)} &Y & -& 1 &1 &{(1)} &N & P& 2 &2 &{(2,2)} &Y & -& 3 &3 &{(3,2,1)} &Y & -&2&1 &(1,3)&N&G&3&3&(3,2,1)&Y&-\\
10 & 3 &3 &{(2,1,1)} &Y & -& 1 &1 &{(1)} &N & P& 3 &3 &{(1,2,1)} &Y & -& 4 &4 &{(2,2,1,1)} &Y & -&3&3 &(1,2,1)&Y & -&4&3&(2,2,1)&N&G\\
11 & 0 &0 &- &N & P& 0 &0 &- &N & P& 0 &0 &- &N & P& 0 &0 &- &N & P&1&0 & -&N&G&4&4&(1,3,1,1)&Y&-\\
12 & 3 &3 &{(1,2,1)} &Y & -& 3 &3 &{(3,2,1)} &Y & -& 3 &3 &{(2,1,1)} &Y & -& 3 &2 &{(2,1)} &N&G&3&3&(1,2,1)&Y & -&1&1&(1)&N&P\\
13 & 1 &1 &{(1)} &N & P& 3 &3 &{(1,3,2)} &Y & -& 2 &2 &{(2,2)} &Y & -& 1 &1 &{(2)} &N & P&3&3&(1,2,1)&Y & -&3&3&(2,2,2)&Y&-\\
14 & 0 &0 &- &N & P& 1 &1 &{(3)} &N & P& 2 &1 &{(2)} &N&G& 0 &0 &- &N & P&2&2 &(3,1)&Y & -&4&4&(1,1,3,1)&Y&-\\
15 & 3 &3 &{(2,1,1)} &Y & -& 2 &2 &{(1,2)} &N & P& 3 &3 &{(2,1,1)} &Y & -& 4 &4 &{(2,2,1,1)} &Y & -&2&2&(2,2)&Y & -&3&3&(3,1,2)&Y&-\\
16 & 3 &3 &{(1,2,1)} &Y & -& 4 &4 &{(2,2,1,1)} &Y & -& 2 &2 &{(2,1)} &N & P& 4 &4 &{(1,2,2,1)} &Y & -&3&3&(1,1,2)&Y & -&3&3&(2,2,2)&Y&-\\
17 & 2 &2 &{(2,2)} &Y & -& 0 &0 &- &N & P& 4 &4 &{(1,1,1,1)} &Y & -& 1 &1 &{(1)} &N & P&3&3&(1,2,1)&Y & -&1&0& -&N&G\\
18 & 0 &0 &- &N & P& 3 &3 &{(2,2,1)} &N & P& 2 &2 &{(1,3)} &Y & -& 3 &3 &{(2,1,3)} &Y & -&2&2&(1,3)&Y & -&3&3&(3,2,1)&Y&-\\
19 & 4 &4 &{(1,1,1,1)} &Y & -& 1 &1 &{(1)} &N & P& 2 &2 &{(1,1)} &N & P& 1 &0 &- &N&G&3&3&(2,1,1)&Y & -&1&0& -&N&G\\
20 & 1 &1 &{(1)} &N & P& 5 &5 &{(1,1,1,1,2)} &Y & -& 2 &2 &{(3,1)} &Y & -& 0 &0 &- &N & P&3&3&(2,1,1)&Y & -&3&3&(2,2,2)&Y&-\\
\midrule
St & \multicolumn{5}{l|}{50\%, 58\%, \pinkbox{1.44}} 
& \multicolumn{5}{l|}{\pinkbox{10\%}, \pinkbox{57\%}, 1.52}
& \multicolumn{5}{l|}{70\%, 85\%, 1.51} 
& \multicolumn{5}{l|}{55\%, 66\%, 1.68}
& \multicolumn{5}{l|}{\limebox{85\%}, \limebox{86\%}, 1.50} 
& \multicolumn{5}{l}{70\%, 77\%, \limebox{1.77}}\\
\bottomrule
\end{tabular}
\begin{tablenotes}
\item[Note 1] \#P -- Number of grouping actions; \#G -- Number of grasping actions; Seq. -- Objects grasped per action; C -- Task completion; St -- Summary.
\item[Note 2] The three values in the St row are, in order: completion rate, delivery rate, and number of objects per grasp.
\end{tablenotes}
\end{threeparttable}}
\end{table*}

\section{Experiments and Analysis}

Our experiments were conducted using a Universal Robots UR3 manipulator with a Robotiq 2F-85 gripper. The target objects were placed within a yellow frame measuring 300 mm $\times$ 400 mm and are expected to be delivered to an adjacent tray. Two Intel RealSense D435 cameras are used as the global and local cameras. Only the RGB data was used and the image sizes were compressed to $240\times320$ pixels to constrain GPU memory consumption. A PC with Intel Core Ultra 7 285K, DDR5-5600 192GB memory, and NVIDIA RTX A6000 GPU was used to train the diffusion policy. When collecting human expert data, another Intel RealSense D435 camera is used to track human hand poses and gripping states. \rev{The primary objects used in the experiments were 3D-printed rectangular cuboids with dimensions of 120 mm $\times$ 20 mm~$\times$ 20 mm.} To evaluate the generalization capability of the learned policy, we additionally prepared a set of shape variants with slightly altered aspect ratios as well as real-world consumer products, such as snack bars. In our particular diffusion policy implementation, we set both the ResNet-18 output dimension $N$ and the positional encoding dimension $P$ to be 128, set $T_o$, $T_a$, and $T_p$ to be 2, 8, and 16, respectively. That is, the learned policy takes the consecutive observations $\boldsymbol{o}_{t-1}$ and $\boldsymbol{o}_{t}$ as input and predicts future actions $\boldsymbol{a}_{t}$, $\boldsymbol{a}_{t+1}$, $...$, $\boldsymbol{a}_{t+7}$ for the next 8 steps. By performing inference every 8 steps and executing the predicted actions in a closed-loop control manner, the system continuously adjusts its behavior. The hidden layers of the U-Net are set to [512, 1024, 2024]. The optimization method used is AdamW, with a learning rate of $10^{-4}$ and a weight decay of $10^{-6}$. The batch size is set to 64. For the diffusion process, the number of denoising steps is set to $K=100$.

\subsection{Experiment 1: Randomly placed 3D printed \rev{cuboids}}

\rev{First, we study the influence of training data scales.} During data collection, we conducted \rev{$300$} demonstrations, where the number of objects was randomly set between $3$ and $6$, and the target objects were randomly placed within the yellow frame. The collected training data were divided into \rev{three} datasets of sizes \rev{100, 200, and 300}. \rev{Separate diffusion policy networks were} trained for each dataset to investigate the effect of data quantity on performance. \rev{Each network} underwent 300000 training steps. The evaluation \rev{was} conducted on tasks with \rev{4 or} 6 objects. The initial UR3 end effector position was set above the bin. $20$ trials were conducted for each number of objects. In each trial, the robot’s performance was assessed until it either successfully completed the task (delivered all objects) or failed. \rev{Detailed data such as the number of grouping actions, grasping actions, the sequence of objects grasped per action, task completion status (i.e., whether all objects were successfully retrieved), failure reasons, and other related metrics were recorded.}
Table \ref{tab_exp} shows the results of the trials and their statistics. The table contains numerous abbreviations. Their meanings are shown in the notes at the bottom. Especially in the statistical section, we computed the following values: (i) Completion rate: The rate at which all objects were successfully delivered over 20 trials. (ii) Delivery rate: The average number of successfully delivered objects divided by the initial number of objects across 20 trials. (iii) \# Objects per grasp: The average number of objects delivered per successful grasp across 20 trials. \rev{These values conceptually reflect} the Overall Success Rate (OSR) metric proposed by Chen et al. \cite{chen2025benchmarkingmultiobjectgrasping}. The maximum values of these statistics across \rev{all} trained networks and numbers of objects are highlighted in lime, while the minimum values are highlighted in pink. \rev{When the value of column C is N, it indicates that the objects were not cleared. In this case, the detailed failure reason is listed in column R, which includes:(i) P: Pushing failure, where objects slipped from the gripper during pushing and failed to be grouped. (ii) G: Grasping failure, where the robot failed to grasp or dropped objects during delivery. (iii) Stagnation: The robot kept stationary or repeated the same action for more than 5 seconds. Stagnation during grouping was classified as a pushing failure, whereas stagnation during delivery was classified as a grasping failure. We can see from the results that as the number of randomly placed graspable objects increases, the task completion rates decrease. Failures frequently occur during pushing and grouping. The frequency of the failures increases with the number of initially randomly placed objects. The majority of pushing failures are caused by stagnation.} This may be attributed to the increasing variety in the input images as the number of objects grows, making it more difficult for the neural network to determine the optimal pushing strategy based on limited information. We can also see from the results that the sequence of grouping and grasping is quite random. The network chose different policies according to object distributions. Furthermore, the results indicate that increasing the training data has a positive impact on all metrics. \rev{Notably, when the training data reaches $300$, the robot grasps an average of 1.77 objects per attempt in the presence of $6$ objects. The robot exhibits improved robustness during pushing and grouping and experienced fewer P failures.} This demonstrates that the policy network can effectively support multi-object grasping and help improve picking efficiency.

Table \ref{tab_timecosts} shows the detailed time costs when applying the policy trained on the dataset trained using \rev{300} demonstrations to trials containing 6 objects. We can see from the results that while multi-object picking involves additional grouping operations that increase the average time per pick, the per-object efficiency improves significantly. In particular, the efficiency (time required per object) is reduced by approximately \rev{8} seconds compared to single-object picking.

\begin{table}[t]
\setlength{\tabcolsep}{2.2pt}
% \captionsetup{labelfont={color=magenta}, textfont={color=magenta}}
\caption{Time costs of 6 objects trained on the 300 dataset}
\label{tab_timecosts}
\rev{
\begin{threeparttable}
\begin{tabular}{c|lllc|c|lllc}
\toprule
ID & $t_1$(s) & $t_2$(s) & $t_3$(s) & T & ID & $t_1$(s) & $t_2$(s) & $t_3$(s) & T\\
\midrule
1 & (12) & (25) & (26) & 63 & 11 & (11,28,13) & - & (26) & 78\\
2 & (32,13) & (30,25) & \rev{-} & 100 & 12 & (30) & - & - & -\\
3 & (17,13) & (18,18) & - & 66 & 13 & - & (12,35,21) & - & 68\\
4 & - & - & - & - & 14 & (12,55,13) & - & (22) & 102\\
5 & - & (25) & - & - & 15 & (13) & (27) & (23) & 63\\
6 & (14,18,14) & - & (25) & 71 & 16 & - & (18,20,22) & - & 60\\
7 & (21) & (24) & (25) & 70 & 17 & - & - & - & -\\
8 & - & (20,22,21) & - & 63 & 18 & (15) & (27) & (23) & 65\\
9 & (13) & (20) & (20) & 53 & 19 & - & - & - & -\\
10 & (29) & (25,22) & - & - & 20 & - & (19,30,23) & - & 72\\
\midrule
\multirow{3}{*}{St} & \multicolumn{4}{l|}{Avg. time for 1 object: ~~19.3 s} & \multicolumn{5}{l}{Efficiency for 1 object: ~~19.3 s}\\
& \multicolumn{4}{l|}{Avg. time for 2 objects: ~22.9 s} & \multicolumn{5}{l}{Efficiency for 2 objects: ~11.4 s}\\
& \multicolumn{4}{l|}{Avg. time for 3 objects: ~23.8 s} & \multicolumn{5}{l}{Efficiency for 3 objects: ~~7.9 s}\\
\bottomrule
\end{tabular}
\begin{tablenotes}
    \item[Note] $t_1$ -- Time required to pick and deliver a single object; $t_3$ -- Time required to pick and deliver two objects simultaneously; $t_3$ -- Time required to pick and deliver three objects simultaneously; $T$ -- Total time required to clear all objects; Efficiency is computed by dividing the average time by the number of simultaneously picked objects.
\end{tablenotes}
\end{threeparttable}
}
\end{table}

\begin{figure}[t]
    \centering
    \includegraphics[width=\linewidth]{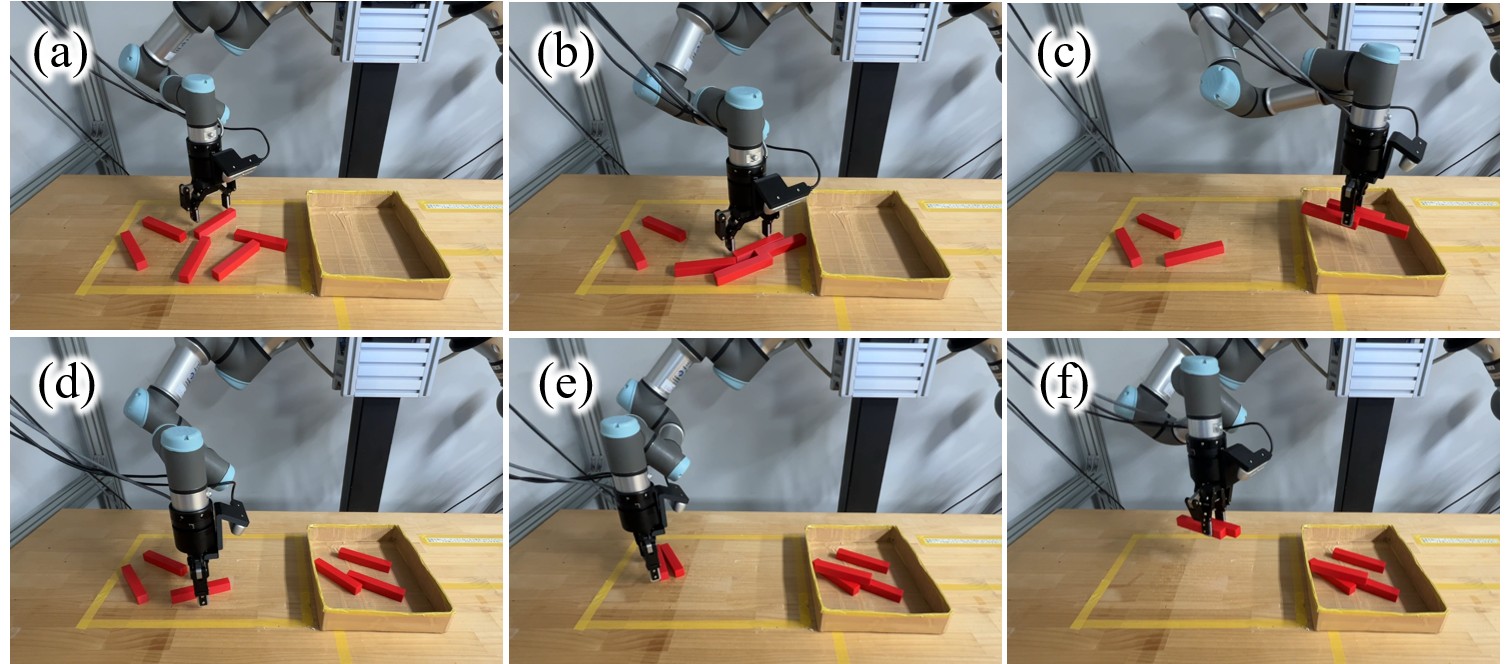}
    \caption{An exemplary result. (a$\sim$c) First grouping and grasping. \rev{(d) Second grasping, (e, f) Third grouping and grasping.}}
    \label{representative}
\end{figure}

Fig. \ref{representative} presents exemplary \rev{snapshots}. The robot first pushed, grouped, and grasped three objects (a$\sim$c), then grouped and grasped two objects (d, e), and finally grasped a single object without pushing (f). The robot selected appropriate policies based on the object arrangement to complete the task successfully. For other detailed robot behaviors and grasping motions, please refer to the supplementary video.

\subsection{\rev{Experiment 2: Comparison with analytical methods}}

\rev{Second, we conducted a comparative study against a rule-based analytical method. Specifically, we improved the approach proposed by Sakamoto et al.~\cite{sakamoto2021efficient} to support combined pushing and grasping for three objects. The method was evaluated under identical experimental conditions with 6 3D printed cuboids and 20 trials.}

\rev{Table \ref{tab_exp2} shows the results. By comparing with Tables \ref{tab_exp} and \ref{tab_timecosts}, we can see that the analytical method achieved similar task completion and object retrieval rates as the model trained using the 300 dataset. However, the proposed method demonstrated significant advantages in terms of time efficiency (1 s inference + 71 s execution vs. 73 s planning + 76 s execution) and achieved a higher grasping efficiency (1.77 vs. 1.44 objects per grasp). Also, a key limitation of the rule-based approach is its reliance on precomputed global motion plans. When minor object displacements occur during pushing motions, subsequent planned grasps become invalid, leading to failures. In contrast, the imitation learning-based approach generates actions sequentially in a closed-loop manner and can adapt online to such disturbances, resulting in improved robustness in cluttered and dynamic scenarios.} \rev{These findings confirm that the policy network provides clear benefits in both efficiency and adaptability over a handcrafted rule-based strategy, especially in multi-object manipulation tasks where object interactions and environmental uncertainties are prevalent.}

\begin{table}[t]
\setlength{\tabcolsep}{3.25pt}
% \captionsetup{labelfont={color=magenta}, textfont={color=magenta}}
\caption{Results of a rule-based method}
\label{tab_exp2}
\centering
\rev{
\begin{threeparttable}
\begin{tabular}{c|clcl|c|clcl}
\toprule
ID & {\#P} & {\#G}-{Seq.} & $t$(s) & C & ID & {\#P} & {\#G}-{Seq.} & $t$(s) & C \\ 
\midrule
 1  & {1} & {0}-{(1)}         & 67+70 & N & 11 & {2} & {2}-{(3,3)} & 42+47 & Y \\  
 2  & {4} & {4}-{(1,1,2,2)}   & 80+96 & Y & 12 & {5} & {5}-{(1,1,1,2,1)}  & 83+73 & Y \\  
 3  & {4} & {4}-{(1,1,2,2)}   & 67+67 & Y & 13 & {3} & {3}-{(1,1,2)}    & 62+36 & N \\   
 4  & {3} & {2}-{(1,2)}       & 54+58 & N & 14 & {3} & {3}-{(1,2,1)}      & 62+33 & N \\   
 5  & {4} & {3}-{(1,1,2)}     & 73+61 & N & 15 & {4} & {4}-{(1,2,1,2)}    & 63+65 & Y \\    
 6  & {5} & {5}-{(1,2,1,1,1)} & 110+94 & Y & 16 & {4} & {4}-{(1,1,2,2)} & 86+70 & Y \\   
 7  & {4} & {4}-{(1,1,2,2)}   & 73+67 & Y & 17 & {2} & {2}-{(1,2)}& 52+37 & N \\ 
 8  & {3} & {3}-{(2,2,2)}     & 55+64 & Y & 18 & {6} & {6}-{(1,1,1,1,1,1)} & 140+128 & Y \\  
 9  & {5} & {5}-{(1,1,2,1,1)} & 69+67 & Y & 19 & {3} & {3}-{(1,2,3)}    & 84+71 & Y \\  
10  & {5} & {5}-{(1,1,1,1,2)} & 83+86 & Y & 20 & {5} & {5}-{(1,1,2,1,1)}  & 59+68 & Y \\ 
\midrule
\multirow{2}{*}{St} & \multicolumn{9}{l}{Completion rate: 70\%; Delivery rate: 87\%; \# Objects per grasp: 1.44;}\\
 & \multicolumn{9}{l}{Average planning time: 73; Average execution time: 76}\\
\bottomrule
\end{tabular}
\begin{tablenotes}
    \item[Note 1] $t$ includes planning time + execution time.
    \item[Note 2] See note of Table \ref{tab_exp} for meanings of abbreviations.
\end{tablenotes}
\end{threeparttable}
}
\end{table}

\subsection{Experiment \rev{3}: Fine-tuning to snack bars}

We conducted fine-tuning on the pre-trained models using \rev{200 and 300 dataset} from the previous experiments and evaluated the adaptivity to real-world commercial products. The fine-tuning dataset consisted of $50$ demonstration samples \rev{using six snack bars, including} three Meiji chocolate bars (size: $125$ mm $\times$ $38$ mm $\times$ $15$ mm, available in two different appearances) and three HiChu fructose bars (size: $130$ mm $\times$ $28$ mm $\times$ $15$ mm, available in three different appearances). The initial object positions were randomly determined, and the fine-tuning process was conducted over $50000$ training steps.

\begin{figure}[!tbp]
    \centering
    \includegraphics[width=\linewidth]{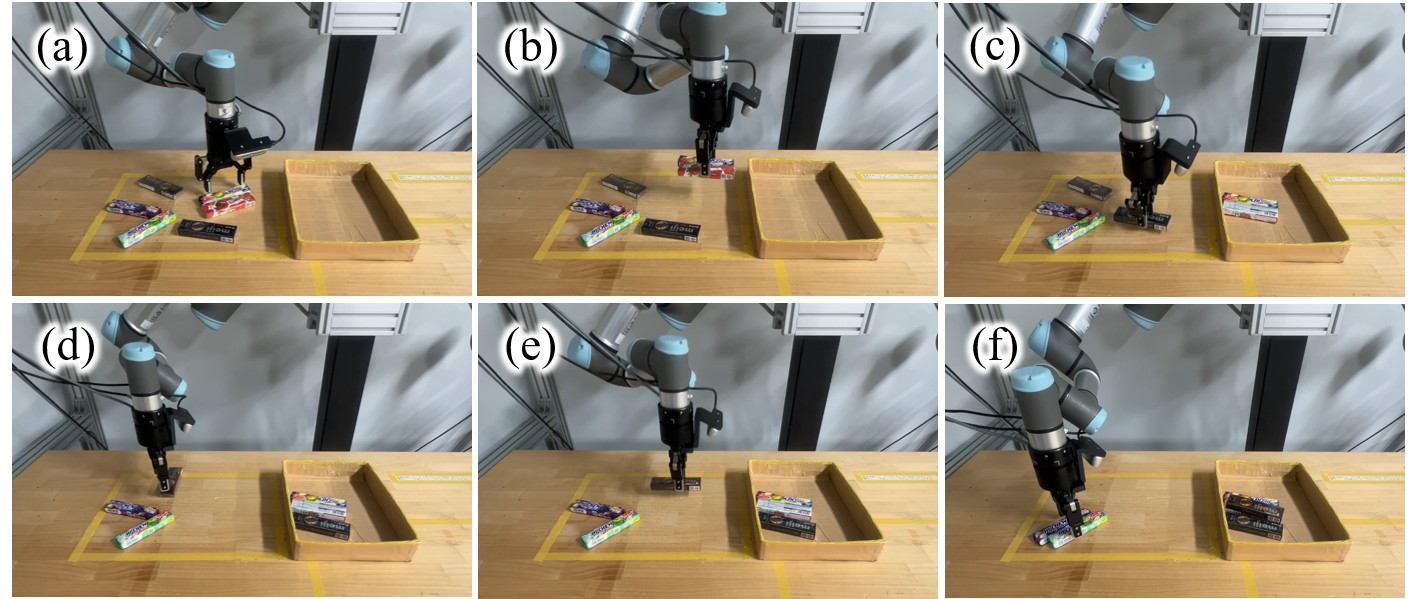}
    \caption{Snack bar trial with a fine-tuned model. \rev{(a$\sim$b) First grouping and grasping. (c) Second grasping. (d, e) Third pushing and grasping. (f) Fourth grouping and grasping.}}
    \label{snackbar}
\end{figure}

Similar to the previous experiments, we carried out $20$ trials for evaluation. Table \ref{tab_exp3} shows the detailed results for these trials. We can see from the results that the task completion rate was 0\% for the 200+50 model and 25\% for the 300+50 model, suggesting that additional learning using a small amount of data is \rev{useful} for transferring policy knowledge effectively \rev{when the foundation training is sufficiently rich}. An important reason for failing to clear all objects was that, during pushing, objects deviated in unexpected directions and became detached from the end effector. Additionally, the robot encountered lots of stagnation and repeated similar policies. Grasping failures also occurred frequently. The stagnation and grasping failures were attributed to differences in the shape and surface texture of the snack bars compared to the objects used in pretraining. \rev{Models with insufficient foundation or limited fine-tuning data struggled to adapt to such differences.} \rev{Meanwhile}, the delivery rate and \# objects per grasp are \rev{respectively 23\% and 1.47 for the 200+50 model and 46\% and 1.31 for the 300+50 model}. They indicate that a certain level of multi-object grasping was achieved. The network can selectively group and grasp a few objects, as shown by the sequence column of the table, the example in Fig. \ref{snackbar}, and also the supplementary video.

\begin{table}[!htbp]
\setlength{\tabcolsep}{2.5pt}
% \captionsetup{labelfont={color=magenta}, textfont={color=magenta}}
\caption{Results of $6$ snack bars using fine-tuned models}
\label{tab_exp3}
\centering
\rev{
\begin{threeparttable}
\begin{tabular}{c|lcc|lcc|c|lcc|lcc}
\toprule
\multirow{2.5}{*}{ID} & \multicolumn{3}{c|}{200+50} & \multicolumn{3}{c|}{300+50} & \multirow{2.5}{*}{ID}
& \multicolumn{3}{c|}{200+50} & \multicolumn{3}{c}{300+50} \\
\cmidrule(lr){2-4} \cmidrule(lr){5-7} \cmidrule(lr){9-11} \cmidrule(lr){12-14}
 & Seq. & C & R & Seq. & C & R &  & Seq. & C & R & Seq. & C & R \\ \midrule
 1  & (2)         & N & P & (2,1,1,1,1) & Y & - & 11 & (2,1,1)   & N & G & (2) & N & G \\
 2  & (2)         & N & P & (2) & N & G & 12 & -         & N & P & (1,2,1,1) & N & P \\
 3  & (2)         & N & G & (2) & N & P & 13 & (1,1)     & N & P & (1,2,2,1) & Y & - \\
 4  & (1)         & N & G & (1) & N & G & 14 & (2)       & N & P & (2,1,1,1,1) & Y & - \\
 5  & (2,1,1)     & N & P & (1) & N & P & 15 & (1)       & N & G & (1) & N & G \\
 6  & (1,1)       & N & G & (1,1,2) & N & P & 16 & -         & N & P & (1,1,1) & N & G \\
 7  & -           & N & P & (2,1) & N & G & 17 & (2,1)     & N & P & (2,1,1,2) & Y & - \\
 8  & (2,2)       & N & G & - & N & G & 18 & (2,1)     & N & G & - & N & G \\
 9  & -           & N & P & (2) & N & G & 19 & -         & N & P & (1,1,2,1,1) & Y & - \\
10  & (1,2,1,1)   & N & G & (1) & N & G & 20 & (1)       & N & P & (1) & N & G \\ \midrule
\multirow{2}{*}{St} & \multicolumn{13}{l}{200+50: Completion: \phantom{0}0\%; Delivery: 23\%; \# Objects per grasp: 1.47} \\
 & \multicolumn{13}{l}{300+50: Completion: 25\%; Delivery: 46\%; \# Objects per grasp: 1.31} \\
\bottomrule
\end{tabular}
\begin{tablenotes}
    \item[Note 1] See note of Table \ref{tab_exp} for meanings of abbreviations.
    \item[Note 2] 200+50 -- A model trained using the 200 rectangular cuboids dataset plus 50 snack bar dataset; 300+50 -- A model trained using the 300 rectangular cuboids dataset plus 50 snack bar dataset;
\end{tablenotes}
\end{threeparttable}}
\end{table}

\subsection{Experiment \rev{4}: Unseen shapes}

We also performed experiments to evaluate the generalization ability of the learned policy across different object geometries. Specifically, we tested the trained policy on L-shaped objects, cubes, and hexagons. The L-shaped objects consist of two connected arms, each measuring 60 mm in length. The cross-sectional edge length is 20 mm. The cube and hexagonal objects both have edge lengths of 20 mm. These dimensions were carefully selected to test the robustness of the policy under \rev{large variations} in object shape and structure.

\rev{Table~\ref{tab_unseen} shows the results without fine-tuning. In this case, the model trained using the 300 dataset was directly applied. Although many grasps were successful, no instances of simultaneous multi-object grasping were observed and none of the trials were fully completed. In all successful grasps, the robot grasped only a single object.} The hexagon shape had the lowest \rev{performance}, likely due to its significant geometric deviation from the training objects. The success rate for the cube \rev{was similarly bad}, as the robot \rev{continuously hesitates around the tube and failure to move forward to grasping}, leading to \rev{stagnation} failures. In contrast, the L-shaped objects achieved \rev{the best performance}. However, some trials involving L-shapes resulted in notably long grasping durations. These delays primarily occurred when one arm of the L-shape was inside the gripper jaws, causing the robot to hesitate between adjusting toward the other arm or attempting to grasp. This ambiguity led to repeated stalling and prolonged execution.

\rev{Table~\ref{tab_unseen_ex} shows the results after fine-tuning with an additional 50 demonstrations tailored to the specific object shapes. As expected, the fine-tuned model exhibited improved performance: despite the small amount of additional data, the robot acquired the ability to group objects based on their spatial configuration and perform simultaneous multi-object grasping. These results suggest that even a modest amount of task-specific adaptation can substantially enhance the model’s ability to generalize and execute more complex manipulation behaviors. Fig. \ref{others} shows some exemplary results of the trials. Full videos can be found in the supplementary file.}

\begin{table}[!htbp]
\setlength{\tabcolsep}{4.7pt}
\centering
\caption{Unseen objects without fine-tuning}
\label{tab_unseen}
\rev{
\begin{threeparttable}
\begin{tabular}{l|clcc|clcc|clcc}
\toprule
\multirow{2.5}{*}{ID} & \multicolumn{4}{c|}{L-shape} & \multicolumn{4}{c|}{Cube} & \multicolumn{4}{c}{Hexagon} \\
\cmidrule{2-5} \cmidrule{6-9} \cmidrule{10-13}
&\#G & Seq. & C & R & \#G & Seq. & C & R & \#G & Seq. & C & R \\
\midrule
1 & 2 & (1,1) & N & P     & 1 & (1) & N & P     & 0 & - & N & P \\
2 & 1 & (1)   & N & P     & 0 & - & N & P       & 0 & - & N & P \\
3 & 2 & (1,1) & N & P     & 2 & (1,1) & N & P   & 0 & - & N & P \\
4 & 4 & (1,1,1,1) & N & P & 0 & - & N & P       & 1 & (1) & N & P \\
5 & 4 & (1,1,1,1) & N & P & 0 & - & N & P       & 1 &  (1) & N & P \\
\bottomrule
\end{tabular}
\begin{tablenotes}
    \item[Note] \#S -- Number of successful grasps; \#F -- Number of failed grasps. All successful grasps involved only a single object. No instances of simultaneous multi-object grasping were observed. The grasping time for each object is shown in the parentheses.
\end{tablenotes}
\end{threeparttable}}
\end{table}

\begin{figure}[!htbp]
    \centering
    \includegraphics[width=\linewidth]{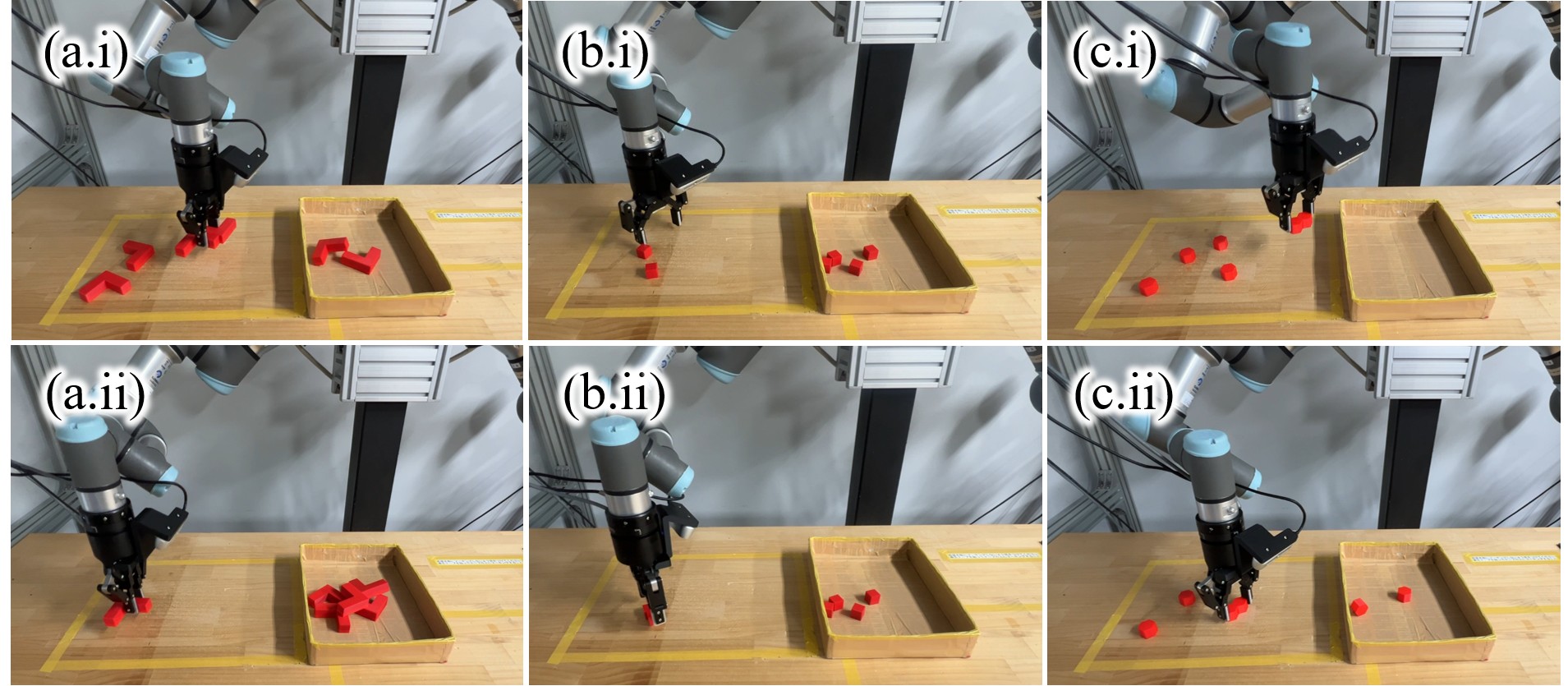}
    \caption{Exemplary results of unseen shapes \rev{using fine-tuned models.} (a.i$\sim$ii) L-shapes. (b.i$\sim$ii) Cubes. (c.i$\sim$ii) Hexagons.}
    \label{others}
\end{figure} 

\begin{table}[!htbp]
\setlength{\tabcolsep}{2pt}
\centering
% \captionsetup{labelfont={color=magenta}, textfont={color=magenta}}
\caption{Unseen objects with fine-tuned models}
\label{tab_unseen_ex}
\rev{
\begin{threeparttable}
\begin{tabular}{l|cclcc|cclcc|cclcc}
\toprule
\multirow{2.5}{*}{ID} & \multicolumn{5}{c|}{L-shape} & \multicolumn{5}{c|}{Cube} & \multicolumn{5}{c}{Hexagon} \\
\cmidrule(lr){2-6} \cmidrule(lr){7-11} \cmidrule(lr){12-16}
& \#P & \#G & Seq. & C & R & \#P & \#G & Seq. & C & R & \#P & \#G & Seq. & C & R \\
\midrule
1 & 0 & 0 & - & N & P           & 3 & 3 & (2,2,1) & N & P   & 4 & 4 & (2,2,1,1) & Y & - \\
2 & 5 & 5 & (1,1,2,1,1) & Y & - & 3 & 3 & (2,2,2) & Y & -   & 1 & 0 & - & N & G \\
3 & 2 & 1 & (1) & N & G         & 2 & 1 & (1) & N & G     & 4 & 4 & (1,2,2,1) & Y & - \\
4 & 3 & 2 & (2,1) & N & G     & 1 & 1 & (2) & N & P       & 4 & 4 & (1,2,2,1) & Y & - \\
5 & 4 & 4 & (2,2,1,1) & Y & -   & 4 & 4 & (2,1,2,1) & Y & - & 1 & 1 & (2) & N & P \\
\midrule
St & \multicolumn{5}{l|}{40\%, 53\%, 1.33} & \multicolumn{5}{l|}{40\%, 67\%, 1.67} & \multicolumn{5}{l}{60\%, 67\%, 1.54}\\
\bottomrule
\end{tabular}
\begin{tablenotes}
    \item[Note] See Table \ref{tab_exp} for meanings of abbreviations and St values.
\end{tablenotes}
\end{threeparttable}}
\end{table} 

\section{Conclusions and Future Work} % .25 page

In this study, we proposed a diffusion policy-based imitation learning framework for grouping and grasping multiple objects. Experiments show that the robot can flexibly perform pushing, grouping, and grasping across various object configurations, and the pre-trained policy can be fine-tuned for new objects. While effective, the method faces limitations when handling large numbers of objects, scaling to larger workspaces, or adapting to diverse shapes and textures. Future work will focus on exploring the impact of expert strategies on data efficiency, improving data collection, and integrating analytical techniques to enhance robustness and generalization.

%%%%%%%%%%%%%%%%%%%%%%%%%%%%%%%%%%%%%%%%%%%%%%%%%%%%%%%%%%%
\normalem
\bibliographystyle{IEEEtran}
\bibliography{citations.bib}
% \bstctlcite{IEEEexample:BSTcontrol}

\end{document}